\documentclass[conference]{IEEEtran}
\IEEEoverridecommandlockouts
\usepackage{cite}
\usepackage{amsmath,amssymb,amsfonts}
\usepackage{algorithmic}
\usepackage{graphicx}
\usepackage{textcomp}
\usepackage{multicol}
\usepackage{multirow}

\def\BibTeX{{\rm B\kern-.05em{\sc i\kern-.025em b}\kern-.08em
    T\kern-.1667em\lower.7ex\hbox{E}\kern-.125emX}}
\begin{document}

\title{Neuromorphic In-Memory Computing Framework using  Memtransistor Cross-bar based Support Vector Machines }

\author{\IEEEauthorblockN{P.Kumar$^\ddagger$$^, $$^\star $, A.R.Nair$^\dagger$$^, $$^\star $, O.Chatterjee$^{\ast \ast}$, T.Paul$^\ast$, A.Ghosh$^\ast$, S. Chakrabartty$^{\ast \ast}$, C.S.Thakur$^\dagger$}\thanks{$^\star $Both the first authors Pratik Kumar and Abhishek Ramdas Nair contributed equally to this paper.}
\IEEEauthorblockA{\{pratikkumar, abhisheknair, tathagata, arindam, csthakur\}@iisc.ac.in, \{oindrila.chatterjee, shantanu\}@wustl.edu \\
$^\dagger$Department of Electronics Systems Engineering, Indian Institute of Science, Bangalore, India, 560012\\
$^\ddagger$Center for Nanoscience and Engineering, Indian Institute of Science, Bangalore, India, 560012\\
$^\ast$Department of Physics, Indian Institute of Science, Bangalore, India, 560012 \\
$^{\ast \ast}$Department of Electrical and Systems Engineering, Washington University in St. Louis,USA, 63130}
}

\maketitle

\begin{abstract}
This paper presents a novel framework for designing support vector machines (SVMs), which does not impose restriction on the SVM kernel to be positive-definite and allows the user to define memory constraint in terms of fixed template vectors. This makes the framework scalable and enables its implementation for low-power, high-density and memory constrained embedded application. An efficient hardware implementation of the same is also discussed, which  utilizes novel low power memtransistor based cross-bar architecture, and is robust to device mismatch and randomness. We used memtransistor measurement data, and showed that the designed SVMs can achieve classification accuracy comparable to traditional SVMs on both synthetic and real-world benchmark datasets. This framework would be beneficial for design of SVM based wake-up systems for internet of things (IoTs) and edge devices where memtransistors can be used to optimize system’s energy-efficiency and perform in-memory matrix-vector multiplication (MVM).

\end{abstract}


\section{Introduction}
\label{sec_intoduction}
\par Due to the proliferation of internet-of-things (IoTs) in the areas of ubiquitous sensing and human-machine interaction, there has been an increased demand towards integrating intelligence directly onto  IoT hardware platforms [1]. In these embedded platforms, high energy-efficiency and low computational/memory foot-print are the key design requirements due to limited battery resources. In this regard, wake-up systems play an integral role and operate by triggering on the computationally and power-intensive modules only when some ambient conditions are detected. As shown in Fig.1, the wake-up system could be a generic signal detector that can sense the input signal and turns on the backend feature extraction and classification module only when the system detects an ambient conditions. It could be a speech signal detector, motion detector in gesture recognition systems, vibration detector in seismic monitoring system and others. Unlike the backend recognition module [2], the wake-up system could have a simpler architecture but must be highly energy-efficient.

\begin{figure}[htbp]
\centerline{\includegraphics[page=1,scale=0.45,trim=4 4 4 4,clip]{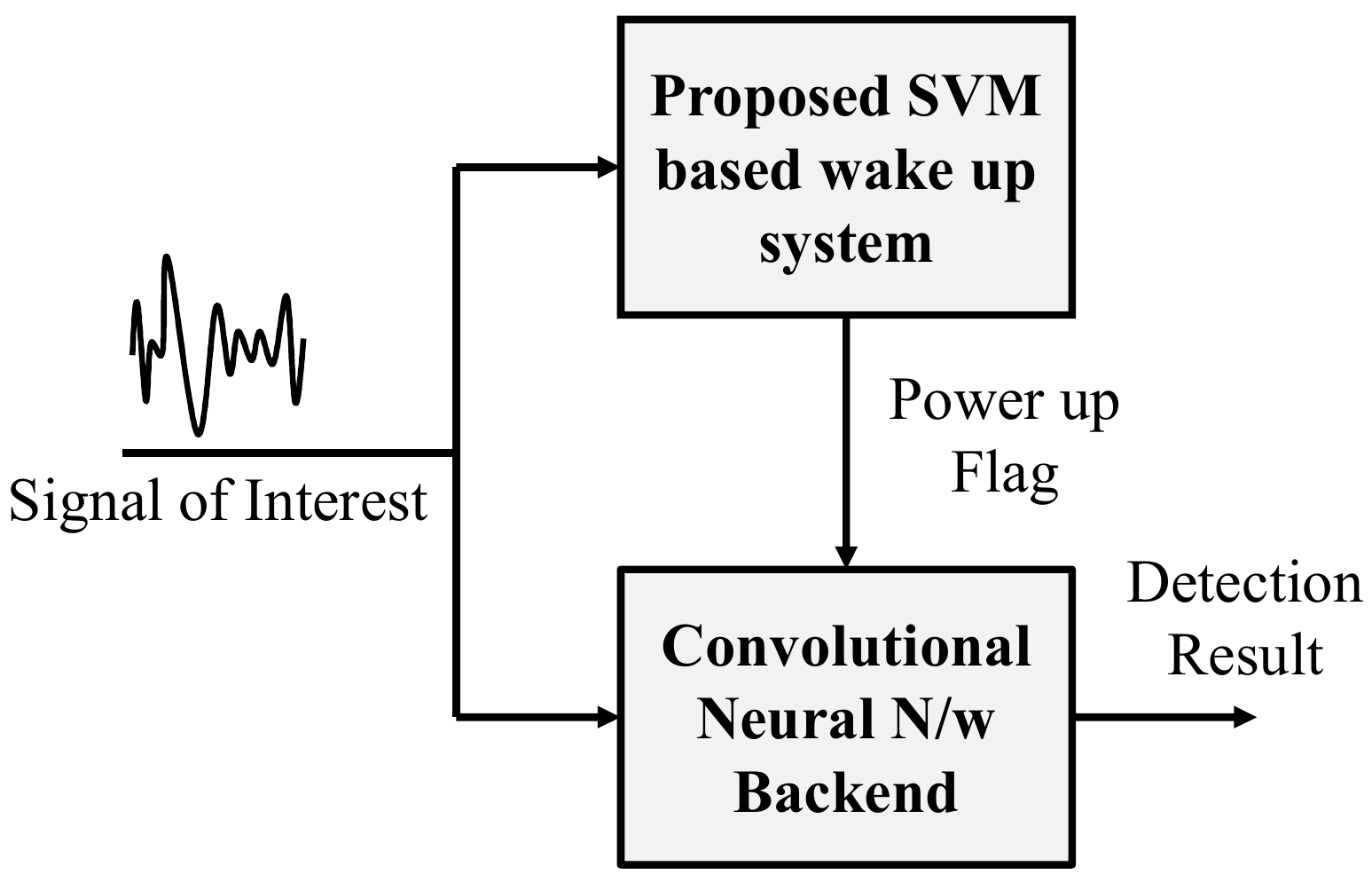}}
\caption{Hybrid system comprising of SVM wake-up for CNN backend.}
\label{fig_motivation}
\end{figure}
  
\par In this scenario, support vector machine (SVM) based wake-up detectors are advantageous because they generalize well with few training samples, their performance is directly determined by its energy-efficiency [3] and under general conditions they offer unique and robust solution. However, evaluating an SVM decision function is computationally intensive [4] since the input features must be matched with several stored templates (or support vectors). Furthermore, SVM architectures require the kernel functions to be positive-definite which leads to uniqueness of the trained solution. To overcome computational complexity, inherent parallelism in SVMs can be mapped onto an array and matrix-based solution for high degree of regularity in computational acceleration [5]. The parallel architecture can also be mapped onto a two-dimensional grid of computing elements interconnected so that shared inputs are along one dimension and shared outputs are along another dimension. Further increase in computational efficiency can be achieved by implementing the array using analog elements where computations such as multiplications are performed using physical properties of devices. In this regard, memtransistor [6] based cross-bar array provides an attractive and energy-efficient platform to implement in-memory computation and matrix-vector operations [7]. The high-density integration offered by nanoscale memtransistor array and it's non-volatility could be exploited to implement SVMs. However, due to the intrinsic non-linearity in memtransistor characteristics, any kernel implemented using its cross-bar array cannot be guaranteed to be positive-definite which is the key requirement for conventional SVMs [4].

\begin{figure*}[ht]
\centerline{\includegraphics[scale=0.54,trim=4 4 4 4,clip]{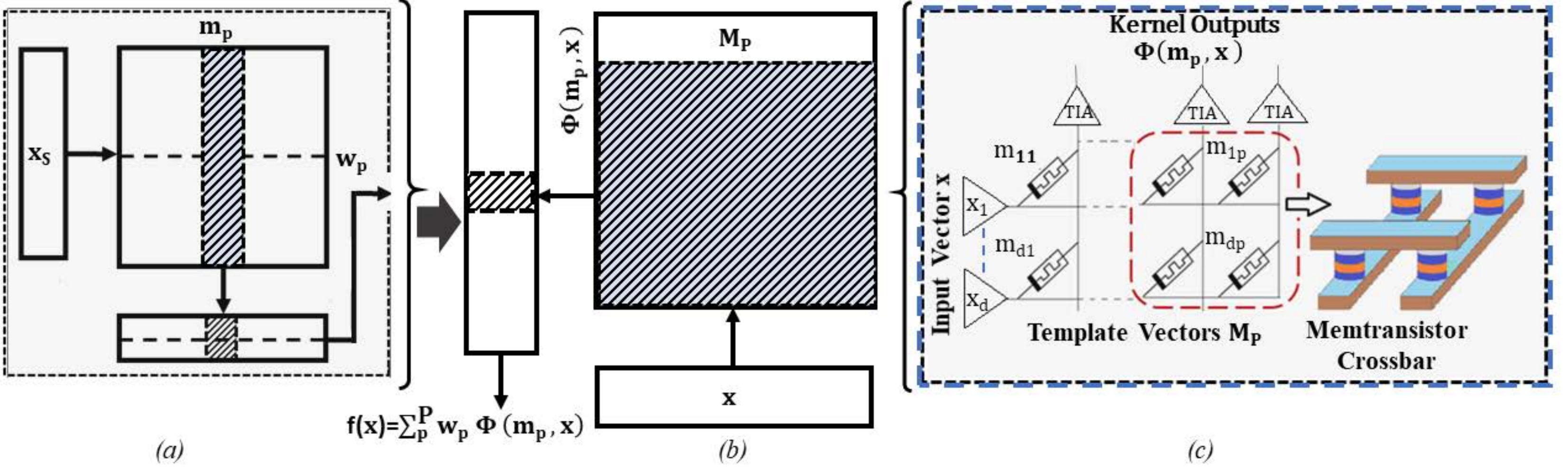}}
\caption{ Schematic of template vector based SVM formulation with memtransistor kernel: (a) Block diagram for computing the similarity of the template vectors with the support vectors; (b) Block diagram for computing the similarity of the input(query) vectors with the template vectors; (c) Memtransistor crossbar array used for kernel computation.}
\label{fig_summary}
\end{figure*}

\par To overcome the need of SVM kernel to be positive definite and to reduce computation complexity along with power requirements, we present a novel framework for designing SVMs. This framework does not impose any explicit restrictions on the nature of the kernel and is robust to fabricated device mismatch and randomness, similar to other work which exploits randomness and is tolerant to device mismatch [8]. Additionally, the fixed number of stored template vectors along with CMOS-Memtransistor cross-bar topology to perform in-memory [9]  computations, significantly improves performance, leading to improved power and area efficiency as compared to traditional SVM based hardware implementations. We used the device measurement data for kernel implementation and showed that the proposed novel framework can achieve state of the art classification accuracy.
 
\section{TEMPLATE BASED SVM FORMULATION}
\label{sec_svm}
\par Given a training set $(\mathbf{x}_i,\mathbf{y_i}),\enskip i=1,\ldots,N$, where $\mathbf{x}_i \in \mathbb{R}^d,\mathbf{y}_i \in \mathbb{R}^c $, the generic form of the decision function for a multiclass SVM is given by $f(\mathbf{x}) = \sum\limits_{s=1}^S \alpha_sK(\mathbf{x}_s,\mathbf{x}) \in \mathbb{R}^c$, where $\mathbf{x}_s,\mathbf{x}_i \in \mathbb{R}^d$ are the ${s}^{th}$ support vector and any arbitrary test vector respectively, $\alpha_s$ is the trained coefficient, $K(\cdot,\cdot)$ is  positive definite kernel, $S$ is the number of support vectors obtained after training and $c$ denotes number of classes in dataset [4,10]. In this paper, we propose a novel variant of the kernel function where instead of computing the similarity of an arbitrary test point with respect to all the support vectors, we precompute the similarity between the support vector and a predetermined set of $P$ template vectors. When a new test point comes in, we compute its similarity only with respect to the template vectors, and synthesize the kernel using the inner product $K(\mathbf{x}_s,\mathbf{x})= \sum \limits_{p=1}^P \Phi(\mathbf{x}_s,\mathbf{m}_p)   \Phi(\mathbf{m}_p,\mathbf{x}),$ where $\Phi(\cdot,\cdot)$ is a non-positive definite function which gives an estimate of the similarity between the ${p}^{th}$ template vector $\mathbf{m}_p$ and the ${i}^{th}$ training vector $\mathbf{x}_i$. The decision function can thus be rewritten as:
\begin{align}
 f(\mathbf{x}) = & \sum\limits_{s=1}^S \alpha_s\sum \limits_{p=1}^P \Phi(\mathbf{x}_s,\mathbf{m}_p)   \Phi(\mathbf{m}_p,\mathbf{x}) \\
 =& \sum \limits_{p=1}^P \Phi(\mathbf{m}_p,\mathbf{x})\sum\limits_{s=1}^S \alpha_s\Phi(\mathbf{x}_s,\mathbf{m}_p)\\
 =& \sum \limits_{p=1}^P \mathbf{w}_p \Phi(\mathbf{m}_p,\mathbf{x}) \end{align}
where $\mathbf{w}_p=\sum \limits_{s=1}^S \alpha_s \Phi(\mathbf{m}_p,\mathbf{x}_s)$ can be thought of as the weight vector obtained after training. Fig.2(a) and (b) show a schematic of the proposed framework, while Fig.2(c) shows a memtransistor based implementation of the same, where the template vectors are programmed into the memtransistor cross-bar as conductances of the memtransistors. The weights are estimated using a standard SVM training procedure using the positive-definite kernels, which ensures a unique solution.
 In our implementation, we have chosen a probabilistic approach [10] for training, even though the framework is also applicable to other SVM training procedures. Input  vector $\mathbf{x}$ is processed by a MVM or a cross-bar array to estimate the kernel functions $\Phi(\mathbf{m}_p,\mathbf{x})$. The kernels are then processed column-wise by the reformulated training weights $\mathbf{w}_p$ and calibrated using the memtransistor crossbar module $\Phi(\cdot,\cdot)$. Fig. 3 shows the flowchart summarizing the design flow for the entire process.  

\begin{figure}[htbp]
\centerline{\includegraphics[page=1,scale=0.45,trim=4 4 4 4,clip]{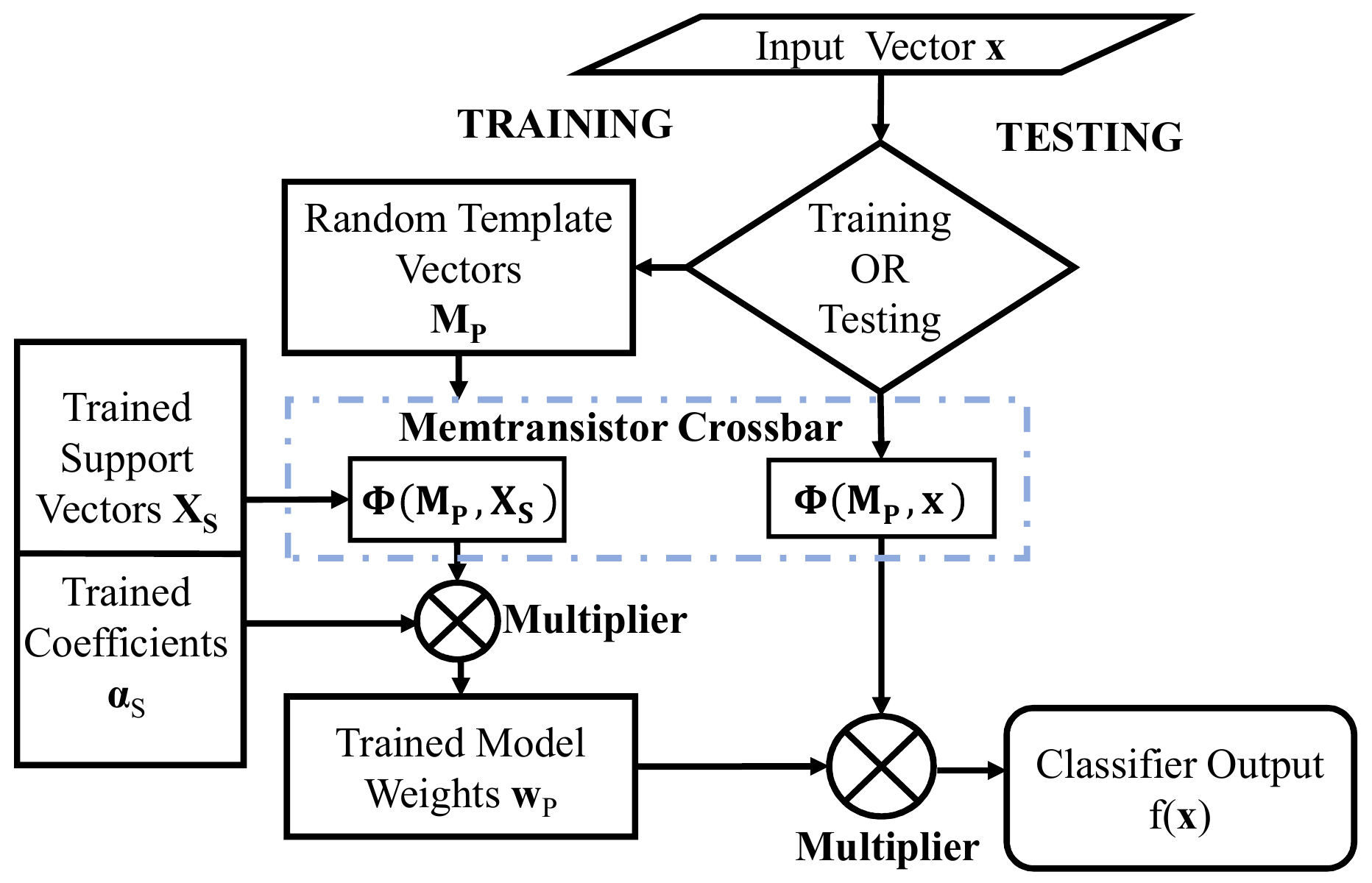}}
\caption{Flow chart describing the off-line training and calibration procedure for the reformulated SVM.}
\label{fig_flowchart}
\end{figure}

\begin{figure*}[htbp]
\centerline{\includegraphics[page=1,scale=0.6,trim=4 4 4 4,clip]{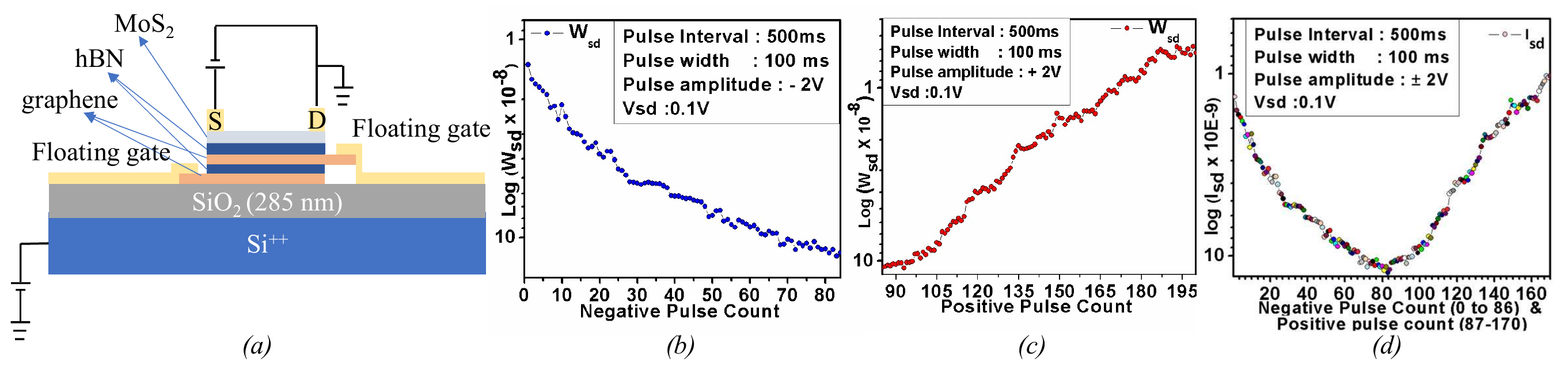}}
\caption{Memtransistor device and characterization: (a) Schematic of fabricated memtransistor device ; (b) Memductance plot for consecutive negative pulse input; (c) Memductance plot for consecutive positive pulse input; (d) Output current vs input voltage(pulse) for negative and positive pulse intervals.}
\label{fig_memtransistor}
\end{figure*}

\section{MEMTRANSISTOR IMPLEMENTATION OF THE TEMPLATE BASED SVM}
\label{sec_memtransistor}
\subsection{Kernel Computation Using Memtransistor Crossbar}

\par Memtransistor crossbar arrays have shown great potential for neuromorphic computing and learning. These non-volatile memories if arranged in crossbar pattern as shown in Fig.2(c) can carry out array size in-memory MVM and addition using Kirchhoff’s current law (KCL) in a single time step. Each column of memtransistor crossbar array represents single template vector with rows equal to the number of features stored as memductance (conductance of memtransistor device). Fig.2(c) shows an ideal crossbar where memtransistors are linear and all other circuit parasitic may be ignored. An input vector voltage $\mathbf{x}$  is applied to the rows of the crossbar, the output voltage  is captured with Trans-Impedance Amplifiers (TIA) and other compensating circuitry at all columns. We get, $\lvert \Phi(\mathbf{m}_p,\mathbf{x})\rvert=\lvert (m_{1p}*x_1)+\ldots+(m_{dp}*x_d)\rvert$ where $\mathbf{M}_P \in \mathbb{R}^{d\times P}$ is the memductance matrix containing all the template vectors. We normalized the input vector between 0 to 1. The memtransistor accuracy (number of repeatable and precise resistance levels) is obtained by the number of memductance states it can maintain and attain. In our case, it can attain around 86 states while maintaining significant readout separation. Energy consumption in fabricated memtransistor was found to be 0.7 nJ for potentiation and 0.5 pJ for depression cycles for the device channel area of $0.423 \times 10^{-14} m^2$, both of which are much lower than traditional CMOS based architectures[11].  

\subsection{Memtransistor Device Fabrication and Characterization}

Fig.4(a) shows a schematic of a typical memtransistor device (symbolic representation used in Fig.3(c)), which consists of a floating gate based two-dimensional multi-state memory device for the storage of weights. The operating principle involves the tunneling of charge carriers from the channel through a tunnel barrier into the floating gate [12]. This charging of the floating gate, in turn, creates an electric field which screens the applied back gate bias leading to a hysteresis in the transfer characteristics and hence memory action. For the proposed application, we have used a memory device which is completely fabricated from ultra-thin two-dimensional layers. The channel of the device constitutes a single layer molybdenum disulphide (MoS$_2$) flake which is semiconducting in nature leading to a high on/off ratio. We have chosen exfoliated hexagonal boron nitride (hBN) as the tunnel barrier because of its single crystalline nature, lack of defects and a large band gap (~5.97 eV). A layer of graphite acts as the floating gate electrode. The device is placed on a Si$ \mathbb{}^+ $$ \mathbb{}^+$/SiO$_2$ (285 nm) wafer, contacts to the sample are made using E-beam lithography followed by packaging in a chip carrier and wire bonding. To obtain the memory action we apply a pulse at the gate( Si$ \mathbb{}^+ $$ \mathbb{}^+$) electrode while simultaneously measuring the conductivity change of the MoS$_2$ channel by applying a very small drain bias (V$_s$$_d$). The memory action is robust and repeatable over multiple switching cycles and the two-dimensional nature of the constituents makes this device geometry immune to short channel effects due to improved gate coupling. We can then combine ‘$n$’ such fabricated devices in crossbar arrangement where each device memductance can be adjusted by applying a pulse of definite width for a definite time. Fig.4(b,c) shows the memductance (M$_{sd}$) variation of memtransistor for the gate pulse of -2 V and +2 V respectively. Fig.4(d) shows the characteristic plot for output current vs input voltage (pulse) obtained for negative and positive pulse intervals. It shows that on applying negative pulses the memductance increases in steps and so does the output current and vice versa for positive pulses.

\section{RESULTS AND DISCUSSIONS}
\label{sec_results}
 We use real physical data obtained from the memtransistor device for mapping the kernel function in order to mimic actual memtransistor crossbar array behaviour. We demonstrated the classification capabilities of the proposed framework for standard datasets and compared it with traditional SVM implementation on both synthetic and real dataset based on multiclass data. Fig. 5(a-c) shows the classification results for three synthetic datasets $100 \times 2$, $100 \times 3$, and $1000 \times 9$ for verification. Tables I and II show a comparison between the classification accuracies of the traditional SVM and the template-vector based SVM on different benchmark UCI datasets such as Stalog Heart, Bank note authentication, Diabetes, Haberman and Activity recognition (AReM) datasets [13]. 
\begin{figure}[htbp]
\centerline{\includegraphics[page=1,scale=0.48,trim=4 4 4 4]{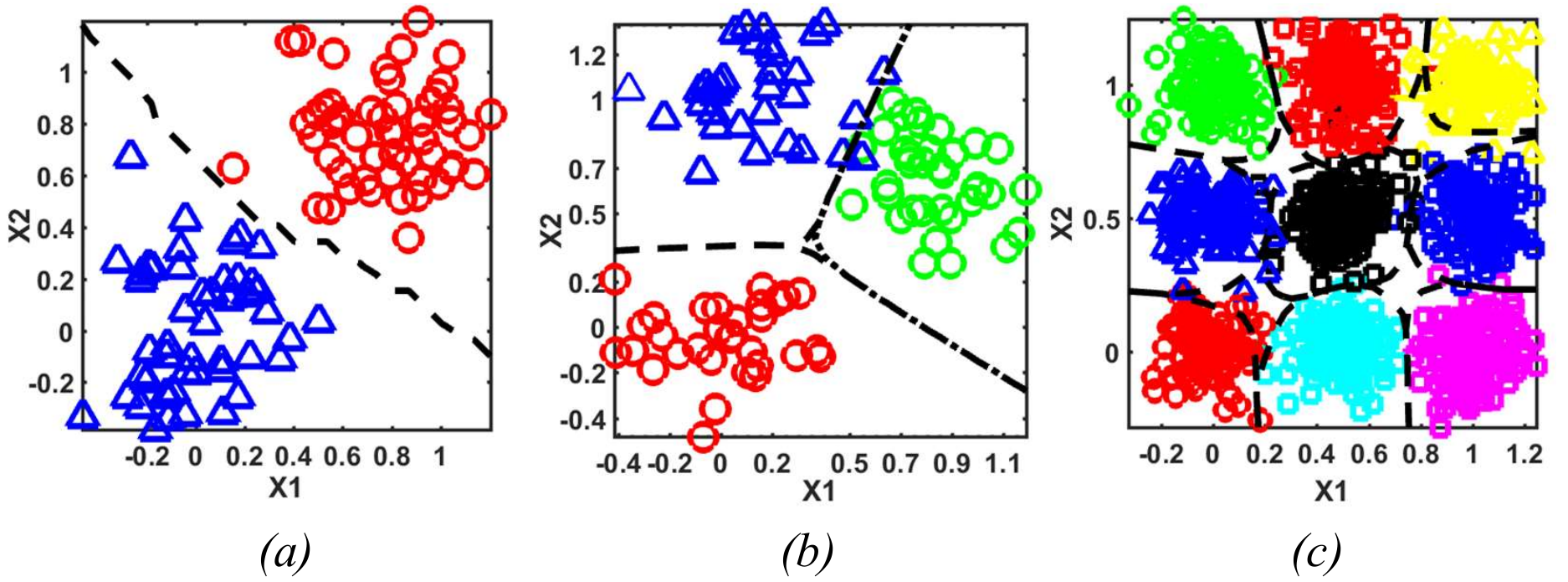}}
\caption{Classification results for the synthetic datasets: (a) Two class; (b) Three class; (c) Nine class datasets.}
\label{fig_results}
\end{figure}
 We combined face datasets from Georgia Tech Face Database (GTFD) [14] and Caltech-101 [15] to generate face and non-face dataset and used it for classification. It can be seen that even with a fixed number of support vectors (here 10 in all the cases except face dataset where 100 support vectors were used) the classification is on par with traditional SVM, yet gives power efficient and memory efficient computations. Furthermore, the support vectors are fixed in number and are implemented using memtransistor whose energy consumption was found to be 0.7 nJ for potentiation and 0.5 pJ for depression cycles for around 90 memory states while occupying a smaller channel area. Additionally, the ability to compute summation by KCL and implementing inherent dot product of memtransistor crossbar provides computational efficiency without any extra hardware.

\begin{table}[!htbp]
\centering
\caption{UCI, GTFD AND CALTECH-101 DATASETS}
\renewcommand{\arraystretch}{1.5}
\begin{tabular}{|c|c|c|c|c|c|c|}\hline
\centering
\multirow{2}{*}{\textbf{Datasets}} & \multicolumn{3}{c|}{\textbf{Traditional SVM}} &\multicolumn{3}{c|}{\textbf{Template-based SVM}} \\
\cline{2-7}& SVs  & Train & Test & SVs & Train & Test  \\ \hline
Statlog Heart & 87 & 87.03 & 81.48 & 10 & 82.69 & 79.25 \\ \hline
Pima Diabetes & 310 & 78.5 & 72.22 & 10 & 76.17 & 73.17 \\ \hline
Bank Note  & 36 & 98.90 & 98.90 & 10 & 88.05 & 88.77 \\ \hline
Haberman & 135 & 75 & 75.8 & 10 & 75 & 71.88 \\ \hline
Face Dataset & 672 & 93.51 & 93.32 & 100 & 95.3 & 86.64 \\ \hline
\end{tabular}
\label{tab_table1}
\end{table}

\begin{table}[!htbp]
\centering
\caption{ACTIVITY RECOGNITION DATASET}
\renewcommand{\arraystretch}{1.5}
\begin{tabular}{|c|c|c|c|c|c|c|}\hline
\centering
\multirow{2}{*}{\textbf{Activity}} & \multicolumn{3}{c|}{\textbf{Traditional SVM}} &\multicolumn{3}{c|}{\textbf{Template-based SVM}} \\
\cline{2-7} & SVs  & Train) & Test & SVs & Train & Test  \\ \hline
Bending & 575 & 95.81 & 96.70 & 10 & 92.74 & 89.61 \\ \hline
Lying & 873 & 93.23 & 91.49 & 10 & 93.46 & 90.99 \\ \hline
Sitting  & 2327 & 79.10 & 78.38 & 10 & 82.18 & 80.12 \\ \hline
Standing & 2589 & 76.86 & 75.43 & 10 & 77.17 & 74.44 \\ \hline
Walking & 1026 & 92.07 & 89.32 & 10 & 90.51 & 89.04 \\ \hline

\end{tabular}
\label{tab_table2}
\end{table}
\section{Conclusions}
\label{sec_conclusions}
In this paper, we presented a unified framework for designing support vector machines (SVMs) that do not impose any explicit restrictions on the kernel to be positive-definite. We also showed that the proposed framework is able to find an SVM solution where the number of stored templates is always fixed. The architecture utilizes our novel memtransistor crossbar topology where the proposed framework itself is robust to fabricated device mismatch and randomness. The measurement data from our fabricated prototype device model was used as fixed stored template vector for memtransistor kernel. Classifications results were presented for both real and synthetic dataset.We also proposed an SVM based wake up system for monitoring incoming signal and deciding whether the signal is relevant for computing task or not such as activity recognition and others. Utilizing only a fixed number of support vector and memtransistor as a kernel, a classification without spending much of the power and memory can be efficiently achieved. Future work includes a implementation of the entire classification framework with a complete memtransistor based memory system for extremely low power envelope and computational cost.

\nocite{*}
\bibliographystyle{ieeetr}

\end{document}